\documentclass[10pt, conference]{IEEEtran}

\usepackage{amsmath}
\usepackage{xspace}
\usepackage{cite}
\usepackage{tabularx}
\usepackage{booktabs}
\usepackage{graphicx}
\usepackage{fontawesome}
\usepackage{tikz,pgfplots}
\usepackage{multirow}
\usepackage{paralist}
\usepackage{flushend}
\usepackage{url}
\usepackage[absolute,overlay]{textpos}

\DeclareMathOperator*{\argmax}{argmax}
\newcommand{\emotionword}[1]{\textsl{#1}\xspace}
\newcommand{\sadness}{\emotionword{sadness}}
\newcommand{\fear}{\emotionword{fear}}
\newcommand{\joy}{\emotionword{joy}}
\newcommand{\anger}{\emotionword{anger}}
\newcommand{\disgust}{\emotionword{disgust}}
\newcommand{\surprise}{\emotionword{surprise}}
\newcommand{\trainbalanced}{\textsc{train}\scalebox{0.8}{\faBalanceScale}\xspace}
\newcommand{\trainall}{\textsc{trainRepr}\xspace}
\newcommand{\testall}{\textsc{testRepr}\xspace}
\newcommand{\ie}{\textit{i.\,e.}\xspace}
\newcommand{\cf}{\textit{cf.}\xspace}
\newcommand{\vs}{\textit{vs.}\xspace}
\newcommand{\eg}{\textit{e.\,g.}\xspace}
\newcommand{\F}{$\text{F}_1$\xspace}
\newcommand{\rt}[1]{\rotatebox{90}{#1}}

\begin{document}
\begin{textblock*}{20cm}(5cm,1cm) 
   2018 International Conference on Data Science and Advanced Analytics
\end{textblock*}
\begin{textblock*}{20cm}(1cm,27cm) 
   \copyright\ 2018 IEEE, preprint of https://doi.org/10.1109/DSAA.2018.00087
\end{textblock*}

\title{An Empirical Analysis of the\\ Role of Amplifiers,
  Downtoners, and Negations\\ in Emotion Classification in Microblogs}

\author{\IEEEauthorblockN{Florian Strohm and Roman Klinger}
\IEEEauthorblockA{Institut f\"ur Maschinelle Sprachverarbeitung\\
University of Stuttgart\\
70569 Stuttgart, Germany\\
Email: \{roman.klinger,florian.strohm\}@ims.uni-stuttgart.de}
}

\maketitle

\begin{abstract}
  The effect of amplifiers, downtoners, and negations has been studied
  in general and particularly in the context of sentiment
  analysis. However, there is only
  limited work which aims at transferring the results and methods to
  discrete classes of emotions, \eg, joy, anger, fear, sadness,
  surprise, and disgust.  For instance, it is not straight-forward to
  interpret which emotion the phrase ``\emph{not happy}''
  expresses. With this paper, we aim at obtaining a better
  understanding of such modifiers in the context of emotion-bearing
  words and their impact on document-level emotion classification,
  namely, microposts on Twitter.
  We select an appropriate scope detection method for modifiers of
  emotion words, incorporate it in a document-level emotion
  classification model as additional bag of words and show that this
  approach improves the performance of emotion classification. In
  addition, we build a term weighting approach based on the different
  modifiers into a lexical model for the analysis of the semantics of
  modifiers and their impact on emotion meaning.
  We show that amplifiers separate emotions expressed with an
  emotion-bearing word more clearly from other secondary
  connotations. Downtoners have the opposite effect. In addition, we
  discuss the meaning of negations of emotion-bearing words. For
  instance we show empirically that ``\emph{not happy}'' is closer to
  sadness than to anger and that fear-expressing words in the scope of
  downtoners often express surprise.
\end{abstract}


\IEEEpeerreviewmaketitle

\section{Introduction}
\label{sec:intro}
Emotion recognition in text is the task of associating words, phrases
or documents with predefined emotions drawn from psychological models
\cite{Ekman1999,Plutchik2001}. In this paper, we phrase it as single
label classification of \joy, \anger, \fear, \sadness, \surprise, and
\disgust.  It has been applied to, \eg, tales \cite{Alm2005}, blogs
\cite{Aman2007}, and as a very popular domain, microblogs on Twitter
\cite{dodds2011temporal}.  The latter in particular provides a large
source of data in the form of user messages \cite{costa2014}, often
used with self-assigned classes by the authors, as this can lead to a
huge albeit noisy data set \cite{Wang2012}. This procedure is often
referred to as \textit{self-labeling}, or, in general, as distant
labeling.

Nowadays, state-of-the-art classification models for emotion
prediction typically take into account sequential information, for
instance with recurrent neural networks or convolutional neural
networks \cite{Felbo2017,Koper2017}. Clearly, these models are able to
capture information expressed in phrases, for instance modifications
of an emotion phrase, like in ``I am slightly unhappy.'' However, such
models do not allow for obtaining a better semantic and linguistic
understanding of the meaning of modifications of emotion expressions
\textit{per se}.

We aim in this paper at getting a better understanding of the impact
and use of modifications of emotion words in Twitter. We perform
modifier cue detection and subsequently identify their scope.
Modifiers are commonly divided into intensifiers (which assign an
intensity to a word) and negators (\eg, \textit{not}), amongst other
classes. Intensifiers are further separated into amplifiers
(\textit{very}, \textit{entirely}, we do not distinguish maximizers
and boosters) and downtoners (\textit{quite}, \textit{slightly})
\cite{Randolph1985}. We focus on these three modifiers:
\textit{negations, amplifiers}, and \textit{downtoners}. From these,
negations are most studied and most challenging in interpretation. For
instance, ``not sad'' might express \joy, \fear, or \anger, or none of
the above. We will argue later that it is closer to expressing \joy
than to \anger or \fear.

Similarly, downtoners might change the prior emotion (\ie, the emotion
of a word or phrase without considering context) of an
expression. However, we will see that for instance ``slightly sad''
most likely still expresses the prior emotion \sadness but also
changes the other emotions which can be expressed by the same sentence
at the same time.
Intensifications (\eg, ``very sad'') seem to be straight-forward in
interpretation. We will argue that such formulations separate the
prior emotion (\sadness) of the word more clearly from a secondary
emotion to be predicted (\eg, \fear).

This research is similar to analyses of the meaning of negations in
the context of sentiment
\cite{Wiegand2010,Kiritchenko2016,Ruppenhofer2015}. However, the
degree of freedom for interpretation is increased due to the greater
set of classes (emotion categories \vs polarity). The only work in the
context of emotions with modifiers we are aware of is by Carillo
\textit{et al.} \cite{Carillo2013}. They focus on the classification
task of sentiment but treat modifiers emotion-specific. In contrast,
we aim at classifying emotions particularly to analyze the role of
modifiers. More specifically, we (1), select and evaluate an
appropriate \emph{modifier scope detection method in the context of
  emotion words} on a manually annotated corpus which we make publicly
available\footnote{The data used in this study is available at
  \url{http://www.ims.uni-stuttgart.de/data/modifieremotion}.}, and
(2), evaluate the impact of the best performing approach in a
\emph{bag-of-words model showing its value for emotion
  classification}. Finally, (3), as the main contribution, we develop
a simple lexical model in which emotion words are weighted differently
based on their modifier scope and prior emotion for the purpose of
\emph{model introspection}: The weights serve as a tool to study the
\emph{meaning of modified emotion words}.

\section{Background and Related Work}
\label{sec:previous}
\subsection{Emotion Analysis}
Ekman defines \joy, \anger, \fear, \sadness, \surprise and \disgust as
the minimal set of six basic emotions that can be differentiated by
facial expressions, the set we use in this paper \cite{Ekman1992}.
Plutchik adds \emotionword{anticipation} and \emotionword{trust} and
the concepts of intensity, emotion mixtures and opposing classes to
the model, which we analyze empirically here \cite{Plutchik2001}.

The first text collection which is nowadays used for emotion
classification is the ISEAR corpus of descriptions of emotional events
\cite{Scherer1997}. Alm \textit{et al.}\ were the first discussing
issues of annotation and prediction of emotions in tales
\cite{Alm2005}.  Aman \textit{et al.}\ built classifiers on top of
blog posts \cite{Aman2007}. Headlines were the subject of analysis in
the SemEval competition on affect recognition \cite{Strapparava2007}.

Next to these manually built corpora, Wang \textit{et al.}\ generated
a training corpus by using the so-called self-labeling information
provided by authors of tweets with their hashtags
\cite{Wang2012}. Their results show that the performance of an emotion
classification system can be significantly improved with a large
amount of data. Similarly, \cite{Purver2012} use self-labeling with
emoticons and hashtags.  The first manually-annotated corpus of tweets
for emotion analysis made publicly available was provided by
\cite{Mohammad2015}, followed by a larger set with a focus on emotion
intensity prediction \cite{Mohammad2017}. The corpus by
\cite{Schuff2017} provides multiple annotations of each instance and
analyzes interactions between classes. It is a re-annotation of a
SemEval corpus for stance detection \cite{Mohammad2016}.

\subsection{Modifier Detection for Sentiment and Opinion Analysis}
\label{sec:mod_detect}
Negations have been extensively studied in different contexts.
Chapman \textit{et al.}\ use a list of negation cue phrases and assume
the scope to include all tokens up to the next punctuation mark or to
the next adversative conjunction \cite{Chapman2001}. Pang \textit{et
  al.}\ include negation detection in a sentiment document
classification system \cite{Pang2002}.

On a more fine-grained level, Councill \textit{et al.} use a lexicon
for negation cue detection and a linear-chain conditional random field
for scope recognition, based on part of speech tags and dependency
relations \cite{Councill2010}. Reitan \textit{et al.} use a similar
approach on a tweet corpus \cite{Reitan2015}. Jia \textit{et al.}\ use
rules based on typed dependencies to determine the scope of a negation
cue \cite{Jia2009}.

A straight-forward approach to modify features in a machine
learning-based text classifier with negation information is to prepend
modified entries in the bag of words (\ie, create an additional bag of
modified words in addition to non-modified words, \eg,
\cite{Pang2002}). For a word-list-based classifier, Polanyi \textit{et
  al.}\ propose to classify a document as positive or negative based
on the sum of weights of positive and negative words
\cite{Polanyi2006}. Positive words have a weighting of $+2$, while
negative words have a weighting of $-2$.  If a word is negated, its
weight is multiplied with $-1$. If a word is amplified, its weighting
is modified additively (to $+3$ or $-3$, respectively) and, if it is
modified by a downtoner accordingly (to $+1$ or $-1$,
respectively). Kennedy \textit{et al.}\ showed an improvement with
this approach on movie review classification
\cite{kennedy2006sentiment}. Follow-up work investaged the use of
negations and modality in a linguistic experiment and also model
multiple negations in the same expression \cite{Benamara2012}.
Taboada et al. discuss lexicon-based methods for sentiment analysis in
a broader context \cite{Taboada2011}. More recent work developed
machine-learning-based classifiers to detect speculation and negations
particularly for sentiment analysis \cite{Cruz2015}.

We are not aware of any previous work on modifier detection for
emotion expressions with the goal of emotion prediction. However,
Carillo \textit{et al.}\ build a model for sentiment classification in
which they learn weights of modifications for an improved polarity
prediction \cite{Carillo2013}.

For a more comprehensive overview of previous work in negation and
modifier detection in sentiment analysis, we refer to surveys and
reviews previously published
\cite{Wiegand2010,Zhu2014,morante2012sem}.

\section{Methods}
\label{chap:methods}
We first aim at showing empirically that handling emotion words
specifically with negations, amplifications, and downtoners improves
the classification in contrast to a purely word-based model.  We
describe our modifier cue detection methods (Section~\ref{sec:mcd}),
explain the modifier scope detection (Section~\ref{sec:MSD}) and
present a simple bag-of-words based method to evaluate the impact of
modifier detection (Section~\ref{sec:emo_class}).

\subsection{Modifier Cue Detection}
\label{sec:mcd}
We limit ourselves to modifications of emotions, in which the modifier
cue $t$ is explicitly mentioned and build on top of existing modifier
lists of negations (\eg, cannot, never, not), amplifiers (\eg,
extremely, very, lot), and downtoners (\eg, few, less, rarely, some)
and merge them
\cite{wikiintensifier,englishclub,thomson1986practical,Romero2012,benamara2007sentiment,Councill2010}. For
a discussion of implicit emotion detection, we refer the reader to our
recent work on the implicit emotion shared task \cite{Klinger2018x}. We
do not differentiate maximizers and boosters \cite{Ito2003}. To focus
our study to those terms which are predominantly used as modifier
instead of other meanings, we calculate
\[r^t_{\textrm{mod}}=\frac{\#t\textrm{ used as modifier}}{\#\textrm{
    used}}\] with
$\textrm{mod}\in\{\textrm{downtoner},\textrm{amplifier},\textrm{negation}\}$
and $\#$ denoting the count. We estimate this value on a corpus
subsample of 100 tweets for each $t$. We accept $t$ as modifier iff
$r^t_{\textrm{mod}}>0.5$ to ensure the main role of a term to be a
modifier. For instance, we dismissed the amplifier \textit{too}, as it
is used more often in a non modifying context. The resulting
dictionaries have 39 negation terms, 69 amplifier terms and 36
downtoner terms.

\begin{table}[t]
\centering\footnotesize
\caption{Features for modifier scope classification\hspace{\textwidth} (proposed by \cite{Councill2010} except for *).}
\label{features}
\begin{tabularx}{\linewidth}{lX}
\toprule
Feature & \multicolumn{1}{l}{Description}  \\ 
\cmidrule(lr){1-1} \cmidrule(lr){2-2} 
Word & Normalized string of a token. \\
\cmidrule(lr){1-1} \cmidrule(lr){2-2} 
POS & Part of speech of a token. \\
\cmidrule(lr){1-1} \cmidrule(lr){2-2} 
Right Dist. & Token distance to the nearest explicit modifier cue in the sentence to the right of a token. \\
\cmidrule(lr){1-1} \cmidrule(lr){2-2} 
Left Dist. & Token distance to the nearest explicit modifier cue in the sentence to the left of a token. \\
\cmidrule(lr){1-1} \cmidrule(lr){2-2} 
Dep Dist. * & Minimum number of edges that must be traversed in the dependency tree from a token to an explicit modifier cue. \\
\cmidrule(lr){1-1} \cmidrule(lr){2-2} 
Dep1 POS & Part of speech of the the first order parent of a token. \\
\cmidrule(lr){1-1} \cmidrule(lr){2-2} 
Dep1 Dist. & Minimum number of edges that must be traversed in the dependency tree from the first order parent of a token to an explicit modifier cue. \\
\cmidrule(lr){1-1} \cmidrule(lr){2-2} 
Dep2 POS & Part of speech of the second order parent of a token. \\
\cmidrule(lr){1-1} \cmidrule(lr){2-2} 
Dep2 Dist. & Minimum number of edges that must be traversed in the dependency tree from the second order parent of a token to an explicit modifier cue. \\
\bottomrule
\end{tabularx}
\phantom{x}\\[1mm]
\end{table}

\subsection{Modifier Scope Detection}
\label{sec:MSD}
As we are specifically interested in the importance and meaning of
modifiers on emotion terms (and not on other words), we only take them
into account in the predictive models where appropriate. We therefore
compare three approaches for modifier scope detection and select the
best performing one.

\subsubsection{Next-$n$ Heuristic}
\label{subsec:nextn}
As a combination of previous work for modifier handling, we define
maximally $n$ tokens as the scope which follow the cue up to the next
punctuation mark or adversative conjunction
\cite{Pang2002,Chapman2001,Hu2004}.  For example, in the tweet
``Happiness is not a goal; it is a by-product.''
the words ``a'' and ``goal'' would be in
the negation scope (with any $n \geq 2$), but not the words following
the semicolon.

\subsubsection{Dependency Tree Heuristic}
\label{sec:DepTree}
\begin{figure}[t]
  \centering
  \includegraphics[width=\linewidth]{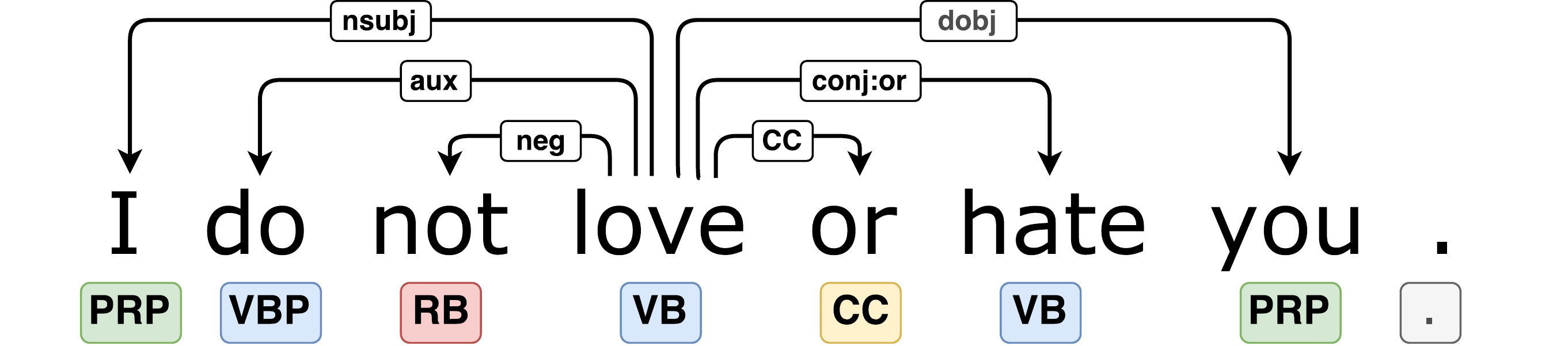}
  \caption{Dependency tree example.}
  \label{dep_tree_1}
\end{figure}

We extend the approach by \cite{Jia2009} to our set of modifiers
and specifically emotion words in a heuristic on dependency trees
(generated with Stanford CoreNLP 3.7.0, \cite{manning2014}): We flag
every parent as modified if its direct child corresponds to a modifier
cue. For instance, in Figure~\ref{dep_tree_1}, ``love'' is recognized
as negated because ``not'' is in our negation lexicon.
To recognize ``hate'' as being in scope as well, we propagate the
modification information along conjunction edges. Adversative
conjunctions block this propagation.

\subsubsection{Binary SVM}
Similarly to a set of submissions to the shared task on negation scope
detection \cite{morante2012sem}, our third approach is a
classification of tokens with linear support vector machines
(SVM). For each modifier, we train one separate model to predict for a
candidate token if it is modified or not.  We assume that a token
cannot be modified twice. The priority of our classifiers is negation
detection, then amplifier detection, followed by downtoner detection.

We use features previously proposed \cite{Councill2010} (\cf
Table~\ref{features}). POS and dependencies are recognized with the
Stanford CoreNLP tools. As an example, the features for the word
``hate'' in Figure~\ref{dep_tree_1} are: \textit{Word = hate},
\textit{POS = VB}, \textit{Right Dist. = 0} (no modifier cue to the
right), \textit{Left Dist. = 3}, \textit{Dep Dist. = 0} (is leaf
node), \textit{Dep1 POS = VB}, \textit{Dep1 Dist. = 1}, \textit{Dep2
  POS = null} (first order parent is root node), \textit{Dep2
  Dist. = 0}.

\subsection{Emotion Classification}
\label{sec:emo_class}
The classification task is to assign a tweet to one of the emotions
from \joy, \anger, \fear, \sadness, \surprise, and \disgust. Note that
we opt for not using a model which can take into account sequential
information (\eg, a long short-term memory, a convolutional neural
net, an $n$-gram model, or non-linear kernels), because the impact of
the modifier detection would be ``hidden'' in the handling of
sequences in general. In contrast, we use a linear support vector
machine with only unigram features such that the SVM is not able to
capture modification effects itself. With this approach we might not
reach highest performance but obtain a model suitable to study
modification effects.

\section{Emotion Classification Experiments under Consideration of Modifiers}
\label{sec:experiments}
In the following Section~\ref{sec:corpora}, we discuss the corpora
used for our evaluation shown in Section \ref{sec:results}, which
shows and discusses the results of our experiments.

\begin{table}
\centering
\caption{Emotion classification corpora.}
\label{emo_corpora}
\begin{tabular}{lrrrrr}
\toprule
emotion & \trainall & \testall & \trainbalanced \\ 
\cmidrule(r){1-1} \cmidrule(lr){2-2} \cmidrule(lr){3-3}\cmidrule(l){4-4}
joy & 597.992& 299.028& 1.000\\
anger & 59.591& 29.501& 1.000\\ 
fear & 68.886& 34.504& 1.000\\
sadness & 207.026& 103.607& 1.000\\
surprise & 24.582& 12.483& 1.000\\
disgust& 1.923& 877& 1.000\\ 
\cmidrule(r){1-1} \cmidrule(lr){2-2} \cmidrule(lr){3-3}\cmidrule(l){4-4}
total & 960.000& 480.000& 6.000\\
\bottomrule
\end{tabular}
\end{table}

\begin{table}
\centering
\caption{Modifier scope detection corpora.}
\label{mod_corpora}
\begin{tabular}{lrrrr}
\toprule
Modifier & \multicolumn{1}{c}{\rt{\textsc{modEval}}} & 
\multicolumn{1}{c}{\rt{\textsc{trainNeg}}} & 
\multicolumn{1}{c}{\rt{\textsc{trainAmp}}} & \multicolumn{1}{c}{\rt{\textsc{trainDown}}} \\ 
\cmidrule(r){1-1} \cmidrule(lr){2-2} \cmidrule(lr){3-3} \cmidrule(lr){4-4} \cmidrule(lr){5-5} 
negation & 315& 630& 0& 0\\
amplifier & 249& 0& 497& 0\\ 
downtoner & 74& 0& 0& 148\\
\cmidrule(r){1-1} \cmidrule(lr){2-2} \cmidrule(lr){3-3} \cmidrule(lr){4-4} \cmidrule(lr){5-5} 
total & 638& 630& 497& 148\\
\bottomrule
\end{tabular}
\end{table}

\subsection{Corpora}
\label{sec:corpora}
\subsubsection{Self-Labeling for Emotion Classification}
To generate corpora of substantial size, we use a self-labeling
approach: we retrieve tweets with specific hashtags for each emotion
using the REST and Streaming APIs provided by Twitter.  The hashtags
are \#glad, \#happiness, \#happy, \#joy, \#lucky, \#luck, and
\#pleasure for \joy, \#anger, \#hate, \#hatred, and \#rage for \anger,
\#afraid, \#angst, \#fear, \#panic, \#scare, and \#worry for \fear,
\#bitter, \#grief, \#misery, \#sad, \#sadness, and \#sorrow for
\sadness, \#surprise and \#surprised for \surprise, and \#disgust for
\disgust.
We assume this hashtags to denote the label of the respective tweets
to create a large dataset. We replace hashtags, URLs, and usernames by
the same strings, respectively.  Table~\ref{emo_corpora} shows our
separation of the crawled data into train and test sets for emotion
classification.  The corpora \trainall and \testall are uniformly
sampled randomly. We use these two corpora to train our emotion
classifier and to evaluate the real world performance and modifier
impact.  Additionally, we create the corpus \trainbalanced which will
be discussed in Section~\ref{sec:emotionlexicon}.

\subsubsection{Manually-annotated Corpora}
\begin{figure}[t]
\centering
  \includegraphics[scale=0.2]{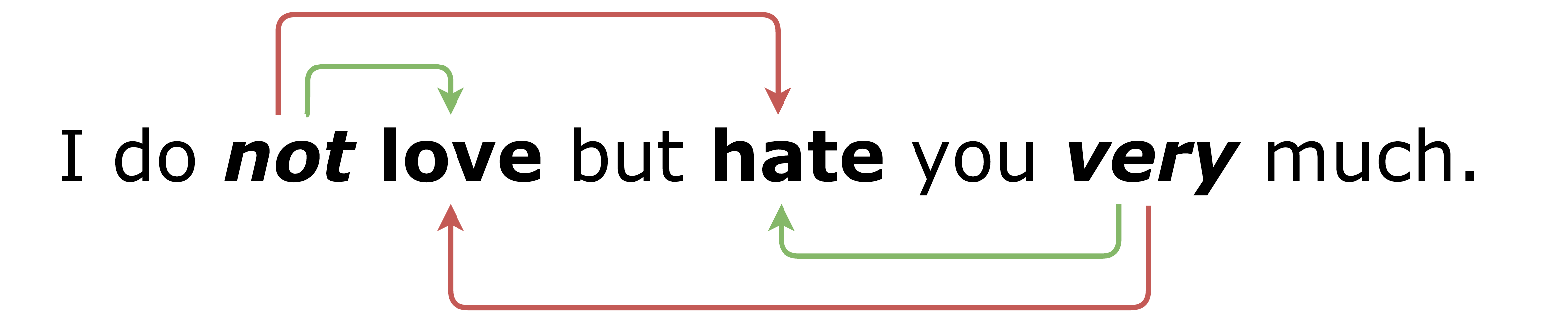}
\caption{Annotation example.}
\label{annotation_example}
\end{figure}

To select the best performing modifier scope detection method and to
estimate their performance, we manually annotate a corpus which is
also used for the SVM scope detection model training. The annotation
is performed by one author of the paper. The task is to categorize
pairs of an emotion-bearing word $z_e$ with a modifier word
$z_{\textrm{mod}}$ into ``$z_{\textrm{mod}}$ modifies $z_e$'' or
``not''.  For instance, Figure~\ref{annotation_example} visualizes
that \textit{not} modifies \textit{love} and \textit{very} modifies
\textit{hate}. However, \textit{not} does not modify \textit{hate} and
\textit{very} does not modify \textit{love}. We therefore have four
instances with two positive and two negative annotations for two
different modifiers and two emotion words.

The resources we create should be valuable also outside of our
specific parameter setting. For instance, our selection of dictionary
entries cannot be complete. Therefore, in the annotation process, the
annotators do not see automatically detected modifiers or
automatically recognized emotion terms but need to mark them
themselves such that the corpus quality is not decreased by error
propagation from preprocessing steps.

Therefore, more specifically, we use three different sampling methods
to obtain a corpus densily populated with relevant instances, but not
limited to those detected with our resources: Equally-sized subsets
are sampled based on the occurrence of (1) both modifier cue and
emotion word, (2) only modifier cue, (3) only emotion word.  Using
different sampling methods enables us to expand our emotion and
modifier lexicons with emotion-bearing words and modifier cues found
during annotation.  We annotate 1,000 tweets resulting in 1,913
modifier-emotion word pairs. Table~\ref{mod_corpora} summarizes the
annotation, split into subcorpora for training the modifier detectors
and an evaluation set: The corpus \textsc{modEval} contains one-third
of the annotations from each modifier type.  We use this corpus to
evaluate the performance of the different modifier scope detection
approaches. Furthermore, we create three corpora \textsc{trainNeg},
\textsc{trainAmp} and \textsc{trainDown} containing the remaining two
thirds of annotations for scope detection model training. The table
also shows that of all detected modifiers, downtoners are the least
common ones.

\begin{table*}[t]
\centering
\caption{Comparison of modifier detection methods on \textsc{modEval}
  corpus. The results of the best method for each modifier and the average are highlighted in boldface
  for precision, recall, and \F, respectively.}
\label{mod_detect_compare}
\begin{tabular}{l|lccccccccc}
\toprule
\multicolumn{2}{l}{}& \multicolumn{3}{c}{Next-2} &
\multicolumn{3}{c}{DepTree} & \multicolumn{3}{c}{SVM} \\
\cmidrule(l){3-5}\cmidrule(rl){6-8}\cmidrule(l){9-11}
\multicolumn{1}{l}{}& Modifier & P & R & F$_{1}$ & P & R & F$_{1}$ & P& R & F$_{1}$ \\
\cmidrule(r){2-2}\cmidrule(lr){3-3}\cmidrule(r){4-4}\cmidrule(l){5-5}
\cmidrule(lr){6-6}\cmidrule(r){7-7}\cmidrule(l){8-8}
\cmidrule(lr){9-9}\cmidrule(r){10-10}\cmidrule(l){11-11}
& Negator & \textbf{93.6}& 87.9& \textbf{90.7}& 93.0& 80.4& 86.2 & 78.7 & \textbf{89.4} & 83.7\\
& Amplifier & \textbf{91.7}& \textbf{93.7}& \textbf{92.7}& 90.7& 83.0& 86.7 & 91.4 &89.4 & 90.4\\ 
& Downtoner& 72.8& \textbf{88.9}& \textbf{80.0}& \textbf{75.0}& 50.0& 60.0 & 66.7 & 55.6 & 60.7\\ 
& Macro-avg. & 86.0& \textbf{90.2}& \textbf{87.8}& \textbf{86.3}& 71.1& 77.7 & 78.9 & 78.2 & 78.3\\
\bottomrule
\end{tabular}
\end{table*}

\subsection{Results}
\label{sec:results}
\subsubsection{Modifier Scope Detection}
\label{sec:modifierscopedetection}
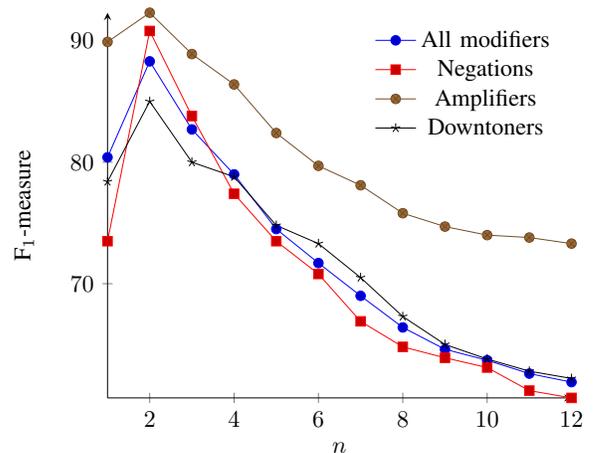
\begin{figure}[t]
  \centering
  \begin{tikzpicture}[scale=0.9]
    \begin{axis}[
      axis lines = left,
      xlabel = $n$,
      ylabel = {\F-measure},
      legend style={draw=none}
      ]
      \addplot+[sharp plot] coordinates
      {(1,80.4) (2,88.3) (3,82.7) (4,79.0) (5,74.5) (6,71.7) (7,69.0) (8,66.4) (9,64.6) (10,63.7) (11,62.6) (12,61.9)};
      \addlegendentry{All modifiers}
      \addplot+[sharp plot] coordinates
      {(1,73.5) (2,90.8) (3,83.8) (4,77.4) (5,73.5) (6,70.8) (7,66.9) (8,64.8) (9,63.9) (10,63.1) (11,61.2) (12,60.6)};
      \addlegendentry{Negations}
      \addplot+[sharp plot] coordinates
      {(1,89.9) (2,92.3) (3,88.9) (4,86.4) (5,82.4) (6,79.7) (7,78.1) (8,75.8) (9,74.7) (10,74.0) (11,73.8) (12,73.3)};
      \addlegendentry{Amplifiers}
      \addplot+[sharp plot] coordinates
      {(1,78.4) (2,85.0) (3,80.0) (4,78.8) (5,74.8) (6,73.3) (7,70.5) (8,67.3) (9,65.0) (10,63.8) (11,62.8) (12,62.2)};
      \addlegendentry{Downtoners}
    \end{axis}
  \end{tikzpicture}
  \caption{Different values of $n$ for next-$n$ modifier
    detection, evaluated on the \textsc{trainNeg}, \textsc{trainAmp},
    \textsc{trainDown} corpora.}
  \label{fig:nextn}
\end{figure}

The results of the selection
of parameter $n$ in the \textit{next-$n$ method}
(Section~\ref{subsec:nextn}) on the training corpora are shown in
Figure~\ref{fig:nextn}. The best result is obtained for $n=2$.
\begin{figure}[t]
  \centering
  \includegraphics[scale=0.8]{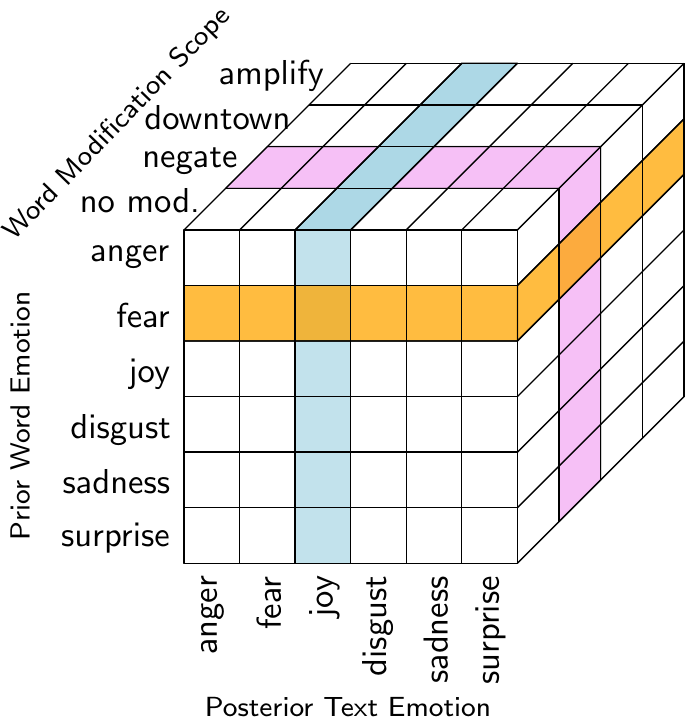}
  \caption{Visualization of our four weight matrices for
    dictionary-based emotion recognition with modifiers. The orange
    slice corresponds to occurrences of input words with prior emotion
    fear, the blue slice to output emotion joy and the red
    slice to input words in the scope of a negation. The cell in all
    three slices contains the weight such word contributes to the
    overall classification.}
  \label{fig:tensor}
\end{figure}
This value goes against our expectations, as \cite{Reitan2015}
detected an average of $n = 3.8$ to work best for negations on Twitter. One reason
might be that we do not consider the full scope of a modifier but limit our analysis to emotion words only.

Table~\ref{mod_detect_compare} shows the results on
\emph{modEval} for emotion scope detection.  The simplest method, the
next-$2$ heuristic, shows best results throughout all modifier
types. The performance for downtoners is substantially lower than for
negations and amplifiers.
The SVM method (83.7\,\% \F for negations) achieves comparable results
to the approach by \cite{Councill2010} (80\,\% \F on product
reviews). The main source of errors for the DepTree approach are errors in the
dependency trees because of missing punctuation.
The source of errors for the comparably low
performance on downtoner scope detection depends on the method.  For
the next-$2$ approach, a challenge is that downtoner cues appear more
often after the scope.  For the DepTree approach, we observe that
downtoner cues are more often not a direct child of associated emotion
words. Regarding the SVM approach, we presume two main reasons for the
limited performance: Firstly, we prioritize negations and amplifiers
and secondly, because we have a limited training set for downtoner.

\begin{table*}[t]
\centering
\caption{Results on \testall corpus and subcorpora limited to instances
  with respective modifiers, with and without modifier detection. The 
  SVM bag-of-words model is trained with unigram features on
  \trainall. The lexicon weight matrices are
  trained on \protect\trainbalanced.}
\label{testAll}
\renewcommand*{\arraystretch}{1}
\setlength\tabcolsep{4.1mm}
\newcolumntype{P}{>{\centering\arraybackslash}p{10mm}}
\begin{tabular}{l|l|r|rrrrrr}
\toprule
\multicolumn{3}{c}{} & \multicolumn{6}{c}{SVM bag of words}\\
\cmidrule(lr){4-9}
\multicolumn{3}{c}{} & \multicolumn{3}{c}{w/o mod.\ det.} &\multicolumn{3}{c}{w/ next2 heur.}\\
\cmidrule(lr){4-6}\cmidrule(lr){7-9}
\multicolumn{1}{c}{}& emotion & size & P&R&\F & P&R&\F\\
\cmidrule(r){1-2}\cmidrule(lr){3-3}\cmidrule(lr){4-6}\cmidrule(lr){7-9}
  \multirow{7}{5mm}{\rt{all data}}
&  joy     & 299,028 & 82.0 & 94.6 & 87.9 & \textbf{83.2} & 93.8 & \textbf{88.2} \\ 
&  anger   & 29,501 & 68.3 & 32.2 & 43.7 & 65.3 & \textbf{34.5} & \textbf{45.1} \\ 
&  fear    & 34,504 & 77.4 & 50.7 & 61.3 & 76.6 & \textbf{53.5} & \textbf{63.0} \\ 
&  sadness & 103,607 & 74.1 & 66.6 & 70.1 & 72.6 & \textbf{68.2} & \textbf{70.3} \\ 
&  surprise& 12,483 & 75.3 & 32.3 & 45.2 & 72.3 & \textbf{33.3} & \textbf{45.6} \\ 
&  disgust & 877 & 18.8 & 3.5 & 5.8 & 17.3 & 3.2 & 5.4 \\ 
&  Macro   & 480,000 & 66.0 & 46.6 & 52.3 & 64.6 & \textbf{47.8} & \textbf{52.9} \\
\cmidrule(r){1-2}\cmidrule(lr){3-3}\cmidrule(lr){4-6}\cmidrule(lr){7-9}
  \multirow{7}{5mm}{\rt{negations}}
& joy		& 22,459 & 70.5 & 83.9 & 76.7 & \textbf{72.3} & \textbf{85.4} & \textbf{78.3} \\ 
& anger		& 5,686 & 61.9 & 35.1 & 44.8 & \textbf{64.3} & \textbf{37.4} & \textbf{47.3} \\ 
& fear		& 6,685 & 75.1 & 50.0 & 60.0 & 70.5 		 & \textbf{57.8} & \textbf{63.5} \\ 
& sadness	& 24,299 & 75.0 & 79.0 & 77.0 & \textbf{77.6} & \textbf{79.1} & \textbf{78.3} \\ 
& surprise	& 1,122 & 39.8 & 12.3 & 18.8 & \textbf{40.7} & \textbf{12.5} & \textbf{19.1} \\ 
& disgust	& 165 & 31.3 & 3.1 & 5.6 & 22.8 		 & 3.1 		 & 5.4 \\ 
& Macro		& 60,416 & 58.9 & 43.9 & 47.2 & 58.1 		 & \textbf{45.9} & \textbf{48.7} \\
\cmidrule(r){1-2}\cmidrule(lr){3-3}\cmidrule(lr){4-6}\cmidrule(lr){7-9}
  \multirow{7}{5mm}{\rt{amplifiers}}
& joy		& 23,622 & 79.6 & 90.8 & 84.9 & \textbf{80.6} & 90.0 & \textbf{85.0} \\ 
& anger		& 3,300 & 61.1 & 29.9 & 40.2 & \textbf{64.0} & 29.2 & 40.1 \\ 
& fear		& 3,017 & 72.6 & 48.7 & 58.3 & 67.9 & \textbf{55.8} & \textbf{61.3} \\ 
& sadness	& 15,773 & 76.9 & 77.0 & 77.0 & 76.1 & \textbf{77.5} & 76.8 \\ 
& surprise	& 872 & 50.4 & 17.5 & 25.9 & \textbf{51.4} & 16.9 & 25.4 \\ 
& disgust	& 109 & 28.6 & 3.7 & 6.6 & 23.9 & \textbf{04.6} & \textbf{7.7} \\ 
& Macro		& 46,693 & 61.5 & 44.6 & 48.8 & 60.7 & \textbf{45.7} & \textbf{49.4} \\
\cmidrule(r){1-2}\cmidrule(lr){3-3}\cmidrule(lr){4-6}\cmidrule(lr){7-9}
  \multirow{7}{5mm}{\rt{downtoners}}
& joy		& 7,900 & 78.2 & 91.1 & 84.1 & \textbf{79.8} & 90.7 & \textbf{84.9} \\ 
& anger		& 979 & 51.9 & 22.0 & 30.9 & \textbf{62.6} & \textbf{35.0} & \textbf{44.9} \\ 
& fear		& 980 & 71.6 & 43.3 & 54.0 & 63.4 & \textbf{48.6} & \textbf{55.0} \\ 
& sadness	& 4,232 & 73.5 & 72.5 & 73.0 & \textbf{74.7} & 72.1 & \textbf{73.4} \\ 
& surprise	& 370 & 54.3 & 15.5 & 24.0 & 46.0 & 15.2 & 22.8 \\ 
& disgust	& 25 & 50.0 & 4.0 & 7.5 & \textbf{66.7} & \textbf{16.0} & \textbf{25.9} \\ 
& Macro		& 14,486 & 63.2 & 41.4 & 45.6 & \textbf{65.5} & \textbf{46.3} & \textbf{51.1} \\
\bottomrule
\end{tabular}
\end{table*}

\subsubsection{Emotion Classification under Consideration of
  Modifiers}
In this paper, we aim at analyzing the impact of negations,
amplifiers, downtoners and to understand their contribution to emotion
analysis, mainly as a justification to further inspect their role on
emotion-bearing words. Therefore, in this result section, we show that
our hypothesis that they affect the interpretation of emotion words
actually holds.

To achieve that, we test our systems on a uniform subsample from
Twitter, namely \trainall/\testall, which has a real-world
distribution of modifiers and non-modified emotions.  For inclusion of
modifier detection, the bag-of-word features of tokens in the scope
are prefixed with respective abbreviations (amp, down, neg) and use
the next-$2$-heuristic.

Table~\ref{testAll} shows the results for SVM classification on four
different subsets of data, namely the full data set for training
(\trainall) and testing (\testall) (called ``all data'' in the table),
the subset of data which contains at least one negator, one
amplifier, or one downtoner, respectively. For these subsets, only
the respective modifier detection is applied.

Altogether, the classifier is best performing on \joy, followed by
\sadness and \fear. The modifier detection contributes consistently,
though partially only to a limited degree, to all class predictions (on
all data for \joy with $+.3$, \anger with $+1.4$, \fear with $+1.7$,
\sadness with $+.2$). Most of the improvement originates from an
increase in recall when training and testing on all data. When
limiting the experiment to different modifiers, we see that this is
likely a result of the negation detection, while amplifiers and
downtoners contribute partially to precision and partially to recall,
depending on the respective emotion.

Inspecting the contribution by modification, we observe the strongest
contribution over the model without handling modifications for
downtoners, with an improvement of $+5.5$ percentage points (pp). Here, 14
pp improvement originate from the emotion \anger and 18 from
\disgust.

Across all modifiers, most important is the special handling of fear,
with 3.5 pp in negations and 3 pp in amplifiers.

\section{Analysis of the Impact of Modifiers in the Context of Emotion
Words}
We showed in the previous section that modifier detection improves
classification in a bag-of-words model. Now we come to the main
contribution of this paper, a deeper analysis of the meaning of
negators, amplifiers, and downtoners on emotion words.

\subsection{Experimental Setting: Weighted Emotion Lexicon}
\label{sec:emotionlexicon}
For this analysis, we extend the work by Polanyi \textit{et al.} from
shifting values in one dimension of polarity according to different
modifiers to multiple dimensions, \ie, six fundamental emotions
\cite{Polanyi2006}. In addition, instead of proposing a fixed set of
weights, we estimate these from data. We use the NRC lexicon for
emotion word recognition, similar to lists of positive/negative words
\cite{Polanyi2006,Mohammad2013}.

The parameters of the model are represented in four $6\times 6$
matrices $W_{\text{no-mod}}$, $W_{\text{amp}}$, $W_{\text{down}}$,
$W_{\text{neg}}$.  In each matrix, one cell $w_{ij}$ corresponds to
the weight which a word of emotion $e_i$ in the respective
modification scope contributes to the emotion $e_j$.  This data
structure is visualized in Figure~\ref{fig:tensor}.
Input text is represented as four count vectors of length 6
($\vec{x}_{\text{no-mod}}$, $\vec{x}_{\text{amp}}$,
$\vec{x}_{\text{down}}$ $\vec{x}_{\text{neg}}$) of words whose scope
contains emotion words of the respective emotion. For instance,
$x_{\text{down},i}$ is the count of downtoned words which belong to
$e_i$.
The posterior emotion score vectors resulting from words of specific
modification scopes for an input text $x$ are then
\[
\vec{e}_{\text{mod}} = W^T_{\text{mod}}\times\vec{x}_{\text{mod}}
\]
with $\text{mod} \in
\{\text{no-mod},\text{amp},\text{down},\text{neg}\}$.
The overall emotion score is then the element-wise sum across rows
\[
\vec{e}(x) = \sum_{\textrm{mod}}\vec{e}_{\text{mod}}
\]
of these vectors.
Finally, the decision for an input text is
\[
e(x) = \argmax_{i} (e_i(x))\,,
\]
where $i$ corresponds to one of the basic emotions.

Based on this setting, we optimize the weights on a balanced corpus
\trainbalanced to further develop an understanding of the meaning of
modifiers by model inspection. The weights are not influenced by
different training set sizes which would make interpretation
difficult. It only includes tweets containing at least one emotion and a
modifier word.  As optimization paradigm, we use hill climbing and \F
as the objective function. We do random restarts with initialization
of $w \sim \mathcal{N}(0,1)$ and take the best matrix from the set of
optimization results. The slice $W_{\text{mod}}$ for each modifier is
optimized for $\approx 120$ hours, resulting in 28 optimization runs
with 2720 epochs on average for the neutral matrix, 49 optimization
runs with 1391 epochs for the negative matrix, 53 optimization runs
with 1248 epochs for the amplifier matrix and 64 optimization runs
with 990 epochs for the downtoner matrix. Weight updates are
performed as $w' = w + r$ with $r \sim \mathcal{N}(0,1)$. We stop
optimization if no improvement is observed in 500 epochs.%
\footnote{We do not report the results of the prediction of this model on
independent data as it is outperformed by the SVM
classification. Instead, we focus on the analysis of the model
parameters in the following.}

\begin{figure*}[t]
\centering
\includegraphics[page=1,scale=0.58]{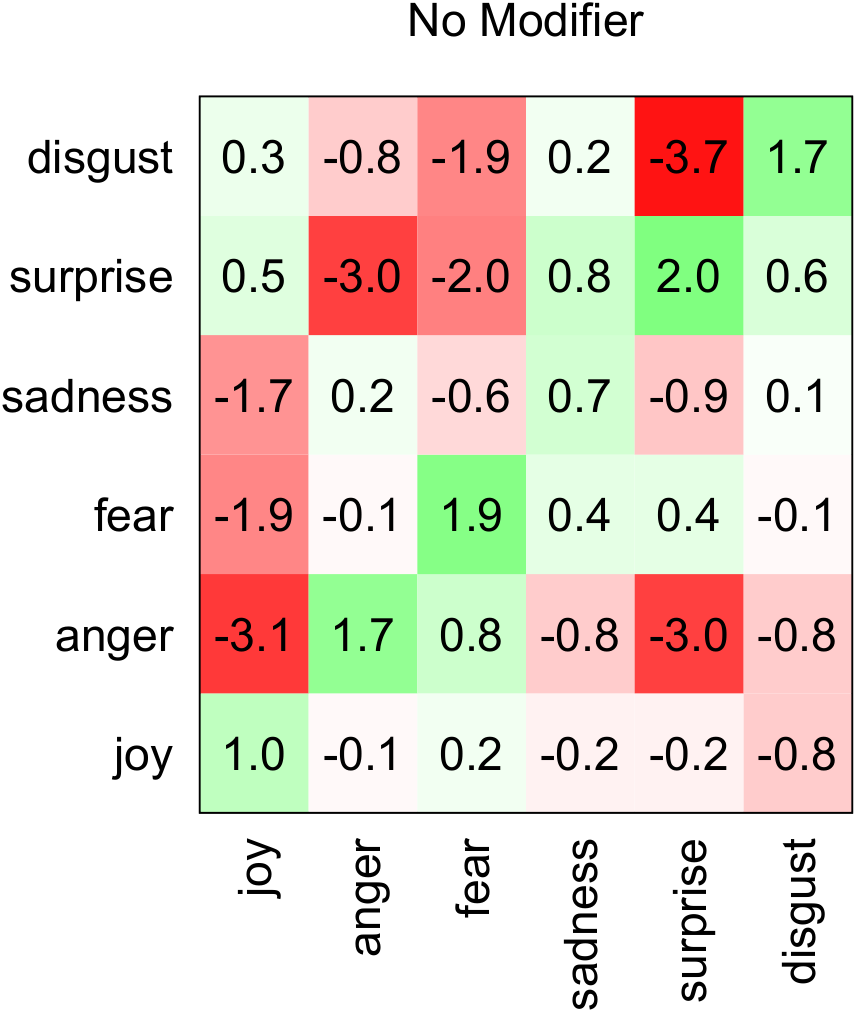}\hfill
\includegraphics[page=2,scale=0.58]{heatmaps-crop}\hfill
\includegraphics[page=3,scale=0.58]{heatmaps-crop}\hfill
\includegraphics[page=4,scale=0.58]{heatmaps-crop}
\caption{Weighting matrices for the lexical model. Columns correspond
  to the predicted emotion, rows correspond to the prior emotion of
  the observed words.}
\label{bestweightingmatrices}
\end{figure*}

\subsection{Analysis of Weighting Matrices in the Lexical Model}
The results of this optimization procedure are shown in
Figure~\ref{bestweightingmatrices}.
We discuss the results based on the following hypotheses:
\emph{Words outside of modifier scope} mainly contribute positively to
the emotion classification corresponding to their prior emotion and
negatively to emotions of opposing polarity.  \emph{Words in negation
  scope} contribute to emotions of their opposing polarity or express
no emotion.  \emph{Words in amplifier scope} contribute more to
emotions of their prior emotion than words outside of modifier scope.
\emph{Words in downtoner scope} contribute less to emotions of their
prior emotion than words outside of modifier scope.

\paragraph{Emotion words outside of modification scope} 
The hypothesis is supported by the matrix; \ie, each word of each
emotion mainly contributes to their prior emotion with the highest
weight for \surprise, followed by \fear, \anger/\disgust, and \joy
(\ie, in a text with \joy and \fear words, both outside of modifier
context, the classification output would be \fear). We observe a
positive contribution of emotion words to other than their prior
emotion for those of same polarity, namely anger to \fear (0.8), \fear
to \sadness (0.4), and \disgust to \sadness\footnote{Example: ``They
  are 'terrorists' not 'Islamists', you {pathetic} excuse for a
  journalist !!!!  \#hate...''} (0.2). Contrasting our expectation,
\surprise contributes to \sadness (0.8), \disgust (0.6), and
\joy\footnote{Example: ``Still can't believe my cute baby shower
  \#afternoontea \#surprise \#ourgirl''} (0.5), showing that \surprise
can be divided into positive and negative realizations.  The negative
contribution of \anger words is striking for \joy ($-3.1$) and
\surprise ($-3.0$), supporting the second of the hypotheses.

\paragraph{Emotion words inside a negation scope}
The hypothesis that negated words mainly contribute to emotions of
opposite polarity holds for \joy to \sadness\footnote{Example: ``Not
  sure how this happened but in two days I've somehow gained 5
  lbs...so {not happy} about this. \#ugly \#fatty \#depressed \#sad''}
(1.2) and \disgust (0.7), and \sadness to \joy\footnote{Example: ``Yes!
  I'm about to eat this piece of cheesecake and I {don't feel guilty}
  about it.  \#indulgingalittle \#cheesecake \#happy''}
(1.0). However, some emotions do not show this flip in polarity, for
instance in \fear and \surprise. For the class \surprise, a
reason is that tweets often use comparisons like the phrase ``\ldots
no party like\ldots'', with ``no party'' indicating negated
\surprise\footnote{Example: ``{Ain't no party like} a birthday party
  when @LJ\_Rader shows up \#surprise''}.  Examples for \fear appearing
in negated context include those whose authors encourage people not to
have fear but still use \fear-related hashtags\footnote{Examples:
  ``Don't worry, let God take control. \#worry'', ``"No fear is
  stronger than you are." - Mark David Gerson \#fear \#quote
  \#spirituality''}.
Altogether, the weights are lower than in the matrix for emotion words
outside of a modifier scope, backing our hypothesis that partially no
emotion is expressed with a negated emotion word.

\paragraph{Emotion words in the scope of amplifiers}
Most diagonal weights of the amplifier matrix show an increased value
in comparison to the matrix for emotion words outside of scope, as
hypothesized (for \joy by factor 2\footnote{Example: ``Wishing you a
  very happy day!  \#happiness \#positivity''}, \fear by factor 1.9,
\sadness by 1.6).  For some emotions, in addition to the hypothesis, the
amplifier clearly strengthens the non-occurrence of another emotion:
an amplified \joy word is a clear signal for non-occurrence of \anger
($-3.8$), while it has nearly no contribution without modification
($-0.1$). This pattern can also be observed for \joy and \fear ($-1.6$
instead of $0.2$ without modification) but only to a lesser degree for
other emotions. For \anger words, the contribution to \sadness flips
from a negative to a positive contribution. Interestingly, amplified
words of \fear contribute positively to all emotions.

\paragraph{Emotion words in the scope of downtoners}
The weights on the diagonal for emotion words in the scope of
downtoners is lower than for words out of scope of a modifier,
however, higher than for negations. Therefore, downtoners can
partially be interpreted as ``light version'' of
negations.\footnote{Example for downtoned sadness with impact on joy:
  ``pray more and {worry less} \#pray \#faith \#love \#peace
  \#happiness...'', and vice versa: ``Just a bit happy to be back in
  Ibiza...''} However, as expected, they do not flip the
polarity. Counter examples are downtoned words associated with \fear
and their impact on \surprise. Most of such tweets contain a phrase
similar to ``little surprise'', which has a meaning similar to
negation. While on average the weights are lower than for other
modifications and no modifications, striking is the highest weight in
all matrices for \sadness contributing negatively to \anger ($-4.3$). A
reason could be that practically no tweet occurs in the corpus that
contains a downtoned word for \sadness and is labeled as \anger.

\section{Summary, Conclusion and Future Work}

In this paper, we showed that modifier detection and handling has an
impact on the prediction of emotions. This impact differs by emotion
and by modifier: the prediction of \disgust and \anger are most
affected by downtoners, while \joy and \anger are most affected by
negations. Amplifications are most relevant to \fear. Across all
emotions the prediction of \surprise and \sadness are not that
strongly affected.

A deeper look on the impact of negations, amplifiers, and downtoners
on separate emotions discloses results which are mostly in line with
the models by Plutchik and Russell \cite{Posner2005}. Interesting
results include that modifiers influence different pairs of emotions
to different degrees: highest weights ($-3.7$) can be observed for
disgust--surprise (\emph{observation}--\emph{prediction}) without
modifiers. Amplifying words denoting \surprise, however, does not
increase such weights but decreases them -- amplifiers separate (some)
emotions stronger from all than their prior emotions. This is
particularly the case for \fear, where the weight increases from 1.9
to 3.6 (without modifier to amplifier). For negations, which are
probably the most challenging modifiers to understand emotions, we see
the highest (negative) weights for \disgust and \fear, \surprise and
\anger -- ``not surprised'' definitely does not mean \anger, and ``not
disgusted'' definitely does not mean \fear. More intuitively are
positive weights which are, again, in line with psychological models.

Future work includes more detailed parameter tuning in our models. We
made the assumption that a maximal \F of scope detection is optimal
for classification and therefore set $n=2$. However, a different ratio
of precision and recall might be beneficial. Therefore, jointly
optimizing parameters of emotion scope detection in the downstream
task might uncover a different parameter setting.

One source of error in the scope detection are mistakes in the parse
tree generation. An evaluation of different parsers and optimizing
them for the task at hand might lead to improved performance.

The weight matrices in our lexical model were optimized separately for
each modifier. However, we represent them as a 3D tensor
already. Therefore, a next step will be a joint optimization of all
parameters. We assume that interactions between them might lead to
improved results.

Our study is built on top of document-level classification. We propose
follow-up studies to investigate the word level and subword level with
the use of distributional semantics. In addition, we did not take into
account implicit modifications and modifying inflections and
derivations. This strain of work will connect our results in this
paper to the initiatives of predicting the intensities of whole
tweets, as shown by Mohammad \textit{et al.}\ in previous work
\cite{Mohammad2017}. In addition, the analysis and comparison with
sequence-based classifiers including attention mechanisms will allow
for a deeper analysis of end-to-end systems. We assume that it is more
challenging to obtain knowledge regarding modifiers from these
methods, however, given the work in this paper, we will analyze if our
hypotheses also manifest in these approaches.

\section*{Acknowledgements}
This research has been partially funded by the German Research Council
(DFG), project SEAT (Structured Multi-Domain Emotion Analysis from
Text, KL 2869/1-1). We thank Evgeny Kim and Laura Bostan for
proof-reading and fruitful discussions.


\end{document}